\documentclass[twocolumn, switch]{article} 

\usepackage{preprint}

\usepackage{amsmath, amsthm, amssymb, amsfonts}
\usepackage{times}
\usepackage{float}
\usepackage{epsfig}
\usepackage{graphicx}
\usepackage{amsmath}
\usepackage{amssymb}
\usepackage{booktabs}
\usepackage{algorithm}
\usepackage{algpseudocode}
\usepackage{tikz}
\usepackage{enumitem}
\setlist[itemize]{noitemsep, nolistsep}
\usetikzlibrary{shapes.multipart, positioning}
\usepackage{graphicx}
\usepackage{subfig}
\usepackage{rotating}
\usepackage{array}

\newcommand\tab[1][1cm]{\hspace*{#1}}

\usepackage[numbers,square]{natbib}
\usepackage[utf8]{inputenc}	
\usepackage[T1]{fontenc}	
\usepackage{xcolor}		
\usepackage[colorlinks = true,
            linkcolor = purple,
            urlcolor  = blue,
            citecolor = cyan,
            anchorcolor = black]{hyperref}	
            
\usepackage{booktabs} 		
\usepackage{nicefrac}		
\usepackage{microtype}		
\usepackage{lineno}		
\usepackage{float}			

\usepackage{lipsum}		
\usepackage{url} 
\usepackage{newfloat}
\DeclareFloatingEnvironment[name={Supplementary Figure}]{suppfigure}
\usepackage{sidecap}
\sidecaptionvpos{figure}{c}

\usepackage{titlesec}
\titlespacing\section{0pt}{12pt plus 3pt minus 3pt}{1pt plus 1pt minus 1pt}
\titlespacing\subsection{0pt}{10pt plus 3pt minus 3pt}{1pt plus 1pt minus 1pt}
\titlespacing\subsubsection{0pt}{8pt plus 3pt minus 3pt}{1pt plus 1pt minus 1pt}

\usepackage{tikz,xcolor,hyperref}

\definecolor{lime}{HTML}{A6CE39}

\title{A Universal Latent Fingerprint Enhancer Using Transformers}

\usepackage{authblk}

\author[1]{André Brasil Vieira Wyzykowski}
\author[2]{Anil K. Jain}

\affil[1]{Department of Computer Science and Engineering, Michigan State University  {\tt\small \{wyzykow2,jain\}@msu.edu}}

\begin{document}

\twocolumn[ 
  \begin{@twocolumnfalse} 
  
\maketitle

\begin{abstract}
Forensic science heavily relies on analyzing latent fingerprints, which are crucial for criminal investigations. However, various challenges, such as background noise, overlapping prints, and contamination, make the identification process difficult. Moreover, limited access to real crime scene and laboratory-generated databases hinders the development of efficient recognition algorithms. This study aims to develop a fast method, which we call ULPrint, to enhance various latent fingerprint types, including those obtained from real crime scenes and laboratory-created samples, to boost fingerprint recognition system performance. In closed-set identification accuracy experiments, the enhanced image was able to improve the performance of the MSU-AFIS from 61.56\% to 75.19\% in the NIST SD27 database, from 67.63\% to 77.02\% in the MSP Latent database, and from 46.90\% to 52.12\% in the NIST SD302 database. Our contributions include (1) the development of a two-step latent fingerprint enhancement method that combines Ridge Segmentation with UNet and Mix Visual Transformer (MiT) SegFormer-B5 encoder architecture, (2) the implementation of multiple dilated convolutions in the UNet architecture to capture intricate, non-local patterns better and enhance ridge segmentation, and (3) the guided blending of the predicted ridge mask with the latent fingerprint. This novel approach, ULPrint, streamlines the enhancement process, addressing challenges across diverse latent fingerprint types to improve forensic investigations and criminal justice outcomes.
\end{abstract}
\vspace{0.35cm}

  \end{@twocolumnfalse} 
] 



\section{Introduction}

Forensic science aims to improve the identification and analysis of latent fingerprints, which are often-invisible residues on surfaces that provide crucial evidence in criminal investigations. Identifying latent fingerprints is challenging due to background noise, overlapping prints, and dust, dirt, or grease contamination. These factors can obscure ridge structures, making differentiating between genuine ridges and artifacts difficult. Furthermore, the lifting process can introduce additional noise and degrade print quality \cite{jain2010fingerprint, 2016Kasper}.

A primary challenge in latent fingerprint analysis is a lower count of minutiae compared to rolled/slap fingerprints, essential for accurate matching \cite{jain2001fingerprint}. The limited fingerprint area captured, often due to partial or smudged impressions, can result in fewer identifiable minutiae. Additionally, latent fingerprints may contain distortions, such as skin deformations, pressure variations, or lateral stretching, complicating the matching process \cite{maltoni2009handbook, bansal2011minutiae}.

\begin{figure}[h]
\centering
    \setlength{\tabcolsep}{1pt}
            \begin{tabular}{ccc}

            \footnotesize NIST SD27 \cite{garris2000nist} & \footnotesize NIST 302 \cite{fiumara2019nist} & \footnotesize MSP Latent \cite{yoon2015longitudinal}\\

            \includegraphics[height=2.65cm]{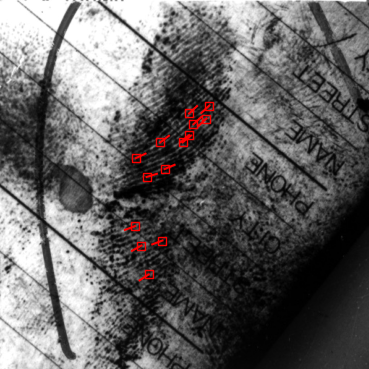}&
            \includegraphics[height=2.65cm]{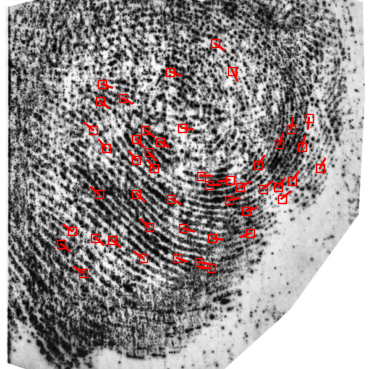}&
            \includegraphics[height=2.65cm]{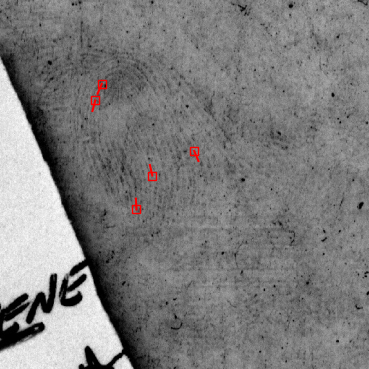}\\
            
            \includegraphics[height=2.65cm]{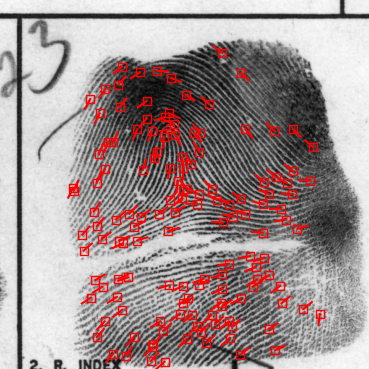}&
            \includegraphics[height=2.65cm]{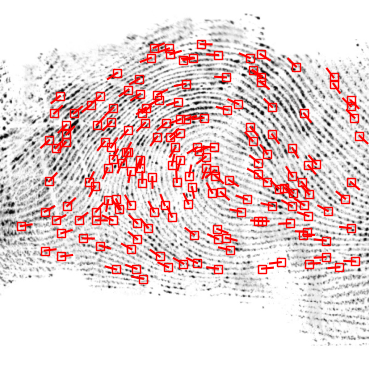}&
            \includegraphics[height=2.65cm]{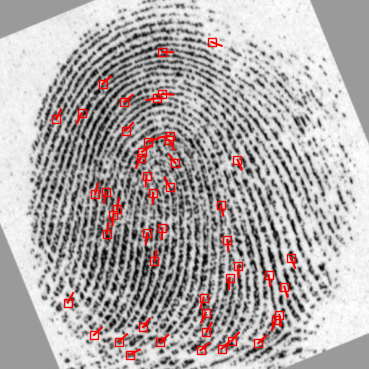}

            \end{tabular}
            \caption{Latent fingerprints (top) and corresponding rolled mates (bottom) with minutiae (red) highlighting differences in the detected minutiae between the two.}
            \label{exampleslatents}
\end{figure}

Figure \ref{exampleslatents} shows several examples of latent fingerprints and their properties that can make matching with rolled fingerprints difficult. 

Fingerprint databases have been created and utilized for research purposes, originating from real crime scenes and others captured under controlled laboratory conditions. Real crime scene databases, such as the NIST SD27 \cite{garris2000nist}, and the MSP Latent database \cite{yoon2015longitudinal}, provide latent fingerprints with all the inherent challenges and variations found in real-world scenarios. However, access to such databases\footnote{NIST retracted NIST SD27 database and MSP database is a private database. NIST 302 (N2N) is a public database collected in a laboratory setting.} is often limited due to concerns about privacy, security, and legal restrictions. On the other hand, laboratory-collected databases, such as NIST SD302 \cite{fiumara2019nist}, NIST SD301 \cite{fiumara2018nist_301} and IIIT-D Simultaneous Latent Fingerprint Database (IIITD-SLF) \cite{yusof2012multi}, are captured under controlled conditions and are typically more accessible for research purposes. While these databases may lack the variability and complexity of real crime scene fingerprints, they are good resources for algorithm development, benchmarking, and testing. 


Current latent fingerprint enhancement methods \cite{yoon2010latent,liban2018latent, huang2020latent} exhibit varying degrees of effectiveness and applicability, with some lacking public implementation or missing crucial details. Furthermore, these methods may not be universally suitable for different latent fingerprint types, resulting in recognition inconsistencies and inaccuracies. Vision Transformer (ViT) \cite{DBLP:journals/corr/abs-2010-11929} has the potential to address these shortcomings and enhance latent fingerprint quality. By leveraging ViT's powerful feature extraction and pattern recognition capabilities, it is possible to develop a more adaptable enhancement method that caters to a wide range of latent fingerprint types. 

Utilizing latent fingerprint databases from laboratory and real-world conditions, a universal latent fingerprint enhancer method addresses challenges by standardizing the enhancement process for diverse fingerprint types. 

Within this context, several factors emphasize the importance of research in latent fingerprints recognition: 

\begin{enumerate}
    \item \textbf{Scarcity of latent databases:} The limited availability of real crime scene fingerprint datasets hampers research progress. Expanding and diversifying databases will give researchers more representative samples, leading to more effective analysis methods for latent fingerprints.
    \item \textbf{Lack of domain experts:}  The multidisciplinary nature of latent fingerprint analysis necessitates skills in chemistry, physics, pattern recognition, and forensic science \cite{sankaran2014latent}. This leads to a scarcity of qualified experts. Furthermore, the number of fingerprints collected at crime scenes often exceeds the capacity of available experts
    \footnote{\href{https://www.bayometric.com/latent-fingerprint-matching-depends-on-human-expertise/}{https://www.bayometric.com/latent-fingerprint-matching-depends-on-human-expertise/}}
    , resulting in potential false positive and false negative errors \cite{ulery2011accuracy}. By improving research in this field, we can develop automated and semi-automated systems to aid experts, making the identification process more accurate. 
    \item \textbf{Bias issues:} Research, including FBI studies, has revealed that latent print examiners are susceptible to confirmation bias, wherein they may inadvertently alter their initial findings due to circular reasoning. To mitigate this, examiners should document their analyses before encountering known fingerprints and record additional information separately. Additionally, research has shown that irrelevant case information can influence examiners' decisions, emphasizing the need for safeguards against exposure to potentially biasing details. This was exemplified in a 2005 terrorism case, where confirmation bias led to incorrect identification and subsequent wrongful detention \footnote{\href{https://obamawhitehouse.archives.gov/sites/default/files/microsites/ostp/PCAST/pcast_forensic_science_report_final.pdf}{https://obamawhitehouse.archives.gov}}. By addressing these biases through continued research, we can enhance the objectivity and reliability of latent fingerprint analysis. 
\end{enumerate}

Our \textbf{research objective} is to develop a method to enhance latent fingerprint quality, including those obtained from real crime scenes and laboratory-created samples, to boost fingerprint recognition system performance.

\textbf{Our contributions are:}
\begin{enumerate}[noitemsep]
    \item A two-step latent fingerprint enhancement method that combines ridge segmentation with UNet \cite{DBLP:journals/corr/RonnebergerFB15} and Mix Visual Transformer (MiT) SegFormer-B5 \cite{DBLP:journals/corr/abs-2105-15203}.
    \item The incorporation of multiple dilated convolutions in the UNet to better capture complex, non-local patterns and enhance ridge and minutiae segmentation.
    \item The application of the Guided Filter algorithm \cite{6319316} to blend the predicted ridge fingerprint mask with the latent fingerprint, preserving ridge and minutiae shape while reducing noise and enhancing details for accurate fingerprint recognition.
\end{enumerate}

Our contributions provide a robust latent fingerprint enhancement method that can be used to streamline forensic investigations, potentially boosting law enforcement agency efficiency.

\section{Background}

Yoon, Feng, and Jain \cite{yoon2010latent} focus on improving the clarity of ridge structures in latent fingerprints using 2D Gabor filters. The algorithm relies on estimating local ridge orientation and frequency, but there are potential shortcomings. For example, inaccurate Gabor filter parameters may weaken accurate ridge structures and inadvertently strengthen spurious ridges. Complex background noise in latent fingerprint images might also negatively affect the orientation field estimation. The approach assumes that examiners can provide accurate manual input, which might be challenging for unclear latent fingerprints. 

Liban and Hilles \cite{liban2018latent} present a methodology for latent fingerprint enhancement, consisting of several steps: normalization, adaptive de-noise using the Edge Direction Total Variation (EDTV) model, reliable orientation image estimation, local frequency estimation, region mask estimation, and Gabor filter application. However, there are potential limitations, such as the method not performing optimally when the latent fingerprint quality is poor or when faced with noise not accounted for by the EDTV model. Additionally, the performance metrics (RMSE and PSNR) may not fully capture fingerprint enhancement quality, as they are global measures and do not specifically evaluate the ridge and minutiae reconstruction.

Huang, Qian and Liu \cite{huang2020latent} tackled the latent fingerprint enhancement problem using a conditional Generative Adversarial Network (cGAN). The proposed model consists of a generator and a discriminator with UNet \cite{DBLP:journals/corr/RonnebergerFB15} and PatchGAN \cite{DBLP:journals/corr/IsolaZZE16} architectures, respectively. They employ a multi-task learning scheme to generate enhanced fingerprint images while outputting orientation field estimation. Despite the promising results, there are limitations, such as the generation of synthetic training data not capturing all variations and noise patterns in real-world latent fingerprints, and the performance metrics used for evaluation may not fully reflect the quality of the enhanced images. 

Despite advancements in fingerprint enhancement techniques, current methods face limitations hindering universal applicability. Jian Li et al. \cite{li2018deep} evaluated their algorithm using limited datasets, obstructing real-world generalization. Techniques like Huang et al.'s cGAN model \cite{huang2020latent} and Liban and Hilles' EDTV model \cite{liban2018latent} demonstrate fingerprint quality improvements but may not fully capture real-world latent fingerprint complexities. Yoon et al.'s 2D Gabor filter-based approach \cite{yoon2010latent} is sensitive to parameter selection and manual input inaccuracies, potentially limiting orientation field model efficacy. Consequently, research is still crucial for a robust latent fingerprint enhancement solution addressing these limitations and promoting generalization in different databases.

\section{Databases}

We incorporated real and synthetic latent fingerprints to ensure diversity and robustness in constructing our training set. We randomly selected a subset of 1097 latent fingerprint images from the MSP Latent database \cite{yoon2015longitudinal} for training and 2000 images from the NIST 302 \cite{fiumara2019nist} for training. The remaining latents within these databases were reserved for subsequent evaluation purposes. The  NIST SD27 database \cite{garris2000nist} was also utilized exclusively for evaluation. Moreover, we have generated 4,000 synthetic latent fingerprints by employing the method developed by Wyzykowski and Jain \cite{Wyzykowski_2023_WACV}, further enriching our training set. A comprehensive overview of the databases utilized in this study is provided in Table \ref{tableDatabases}.

\begin{figure}[H]
\centering
    \setlength{\tabcolsep}{1pt}
            \begin{tabular}{ccc}

            \includegraphics[height=2.53cm]{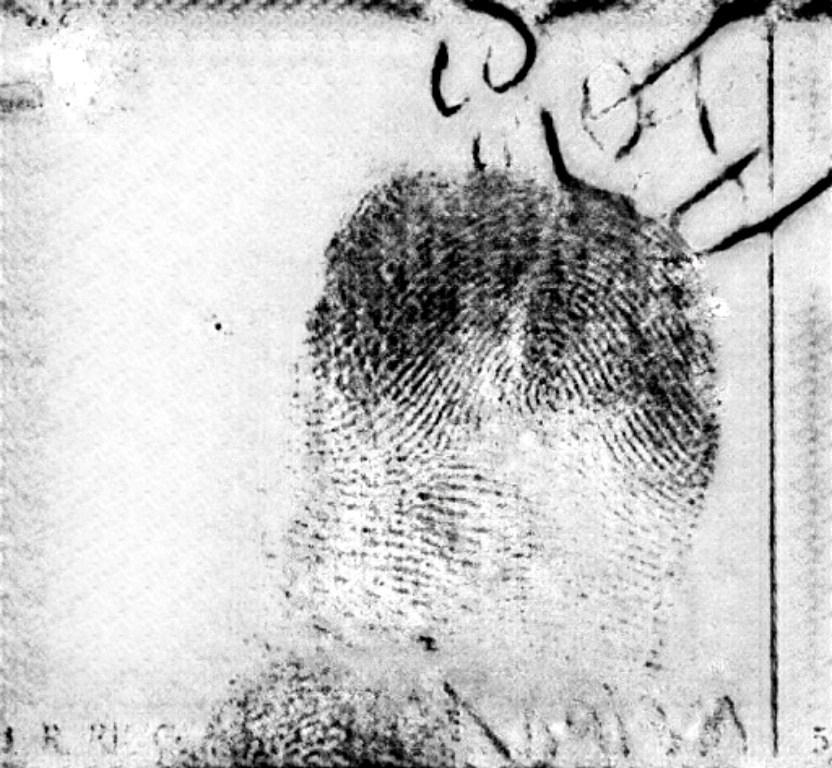}&
            \includegraphics[height=2.53cm]{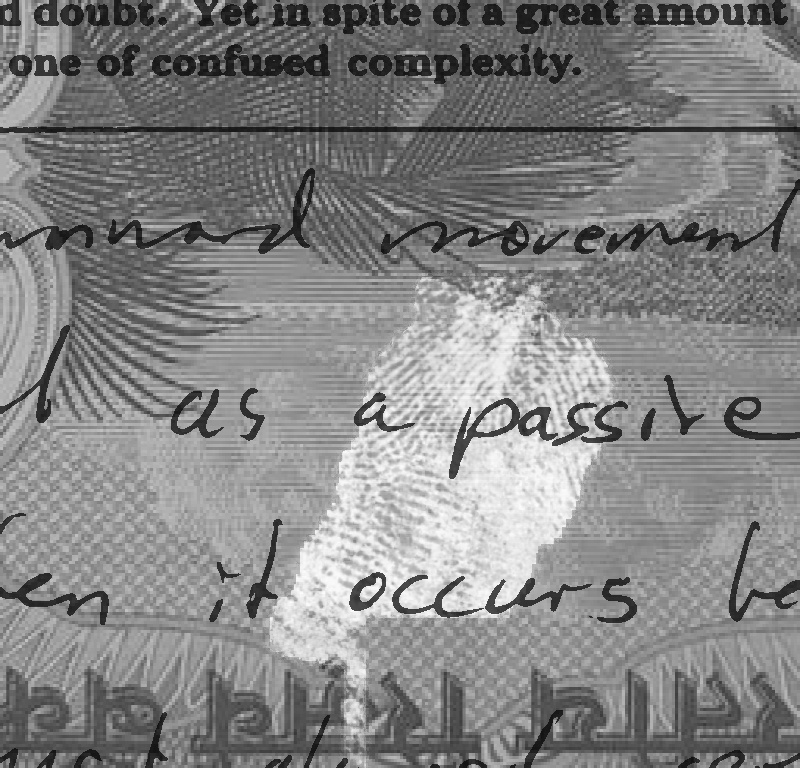}&
            \includegraphics[height=2.53cm]{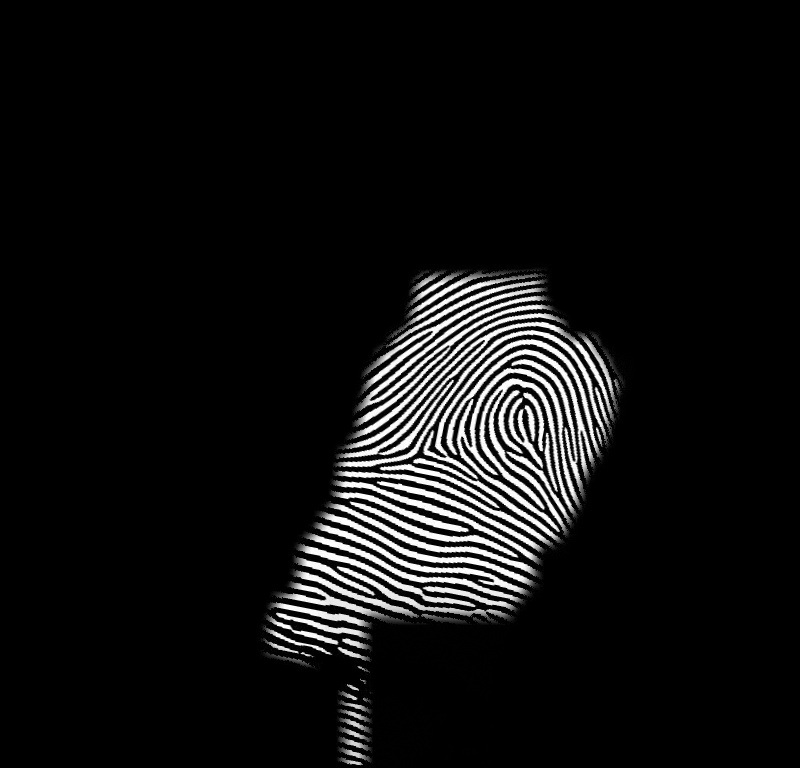}\\

            \footnotesize Original SLP latent & \footnotesize Enhanced SLP & \footnotesize Ridge mask
            
            \end{tabular}
            \caption{Augmenting complex backgrounds in SLP \cite{Wyzykowski_2023_WACV}.}
            \label{slp_exemples}
            \vspace{-0.5cm}
\end{figure}

To enhance the realism of the synthetic latent fingerprints in the SLP \cite{Wyzykowski_2023_WACV} method, we have integrated a variety of backgrounds, including images from banknotes from the Dataset of Indian and Thai Banknotes \cite{cjb5-n039-20}, as well written texts from the IAM-database \cite{marti2002iam}. We applied random translations and rotations to augment the diversity of these synthetic latents. We then extract the ridge mask using Verifinger SDK V12.4. Those ridge masks are used as ground truth in our training. Finally, we perform erosions to the fingerprint mask, reducing the latent size area. Figure \ref{slp_exemples} illustrates the incorporation of complex backgrounds on the SLP synthetic latents.

\begin{table}[H]
\centering
\resizebox{\columnwidth}{!}{%
\begin{tabular}{@{}ccccc@{}}
\toprule
\textbf{Database} &
  \textbf{No. of latents} &
  \begin{tabular}[c]{@{}c@{}}\textbf{No. of images}\\ \textbf{used for training}\end{tabular} &
  \textbf{Source} &
  \textbf{\begin{tabular}[c]{@{}c@{}}Public \\ domain\end{tabular}} \\ \midrule
NIST SD27  & 258   & -     & Crime scene                                                      & No  \\
MSP Latent & 2,074 & 1097  & Crime scene                                                      & No  \\
NIST SD302 & 9,990 & 2,000  & \begin{tabular}[c]{@{}c@{}}Laboratory \\ collection\end{tabular} & Yes \\
SLP \cite{Wyzykowski_2023_WACV}        & 4,000 & 4,000 & Synthetic                                                        &  No    \\ \bottomrule
\end{tabular}
}
\caption{Summary of latent fingerprint databases}
\label{tableDatabases}
\footnotesize{*The SD302h subset contains 9,990 latent fingerprints, obtained after filtering matched rolled and latents from finger position annotations.}
\vspace{-0.3cm}
\end{table}

\begin{figure*}[h]
   \centering
     \includegraphics[height=4.1cm]{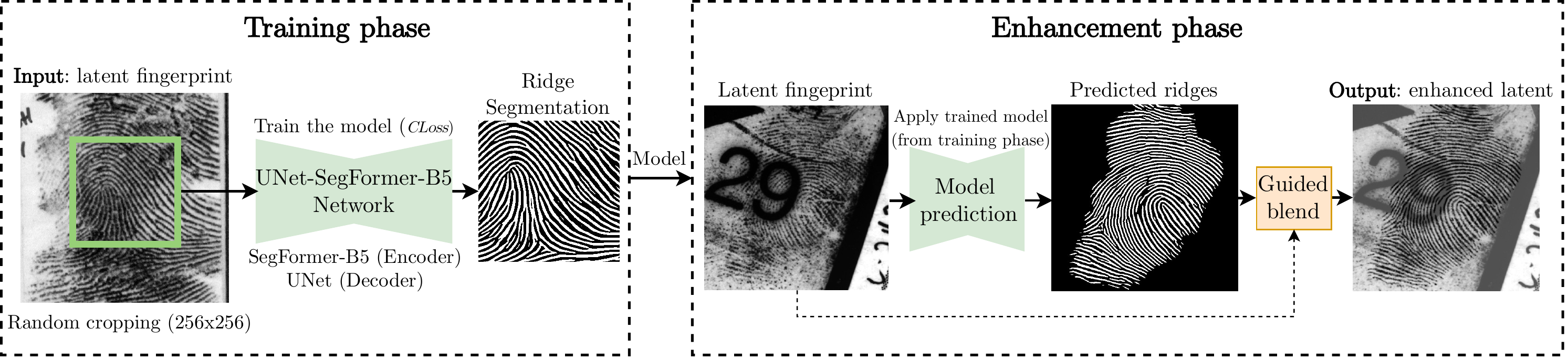}
    \caption{Steps to enhance latent fingerprints using the proposed approach.}
    \label{Flowchart}
    \vspace{-0.3cm}
\end{figure*}

\section{Training Set Creation}

Despite incorporating augmented synthetic SLP latents with various background complexity into our training, using real latents remains crucial for effective ridge and minutiae segmentation. Consequently, we developed a pre-segmentation method to generate ground truth data with segmented ridges.

The accurate segmentation of fingerprint ridges is essential for reliable latent identification. The challenge of achieving robust segmentation is compounded by the inherent complexity and diversity of latent fingerprint patterns and the variability in the quality of latent fingerprints, often contaminated by noise, distortions, and partial impressions. A robust training set is indispensable for creating a ridge segmentation model.

To achieve this, we created a pipeline incorporating real latent fingerprints, creating the ridge ground truth of real latents, which we detail in Section \ref{trainin1}. 

\subsection{Ridge mask training set creation}
\label{trainin1}

Human annotation of latent print images is challenging due to time constraints, required expertise, and annotator subjectivity. Complications arise from low visibility, complex patterns, image distortion, and inconsistent print quality. Incomplete prints and human errors affect reliability and accuracy, while mental exhaustion reduces effectiveness. To address these issues, we created an automatic solution to extract ridge masks for the training set of real latents from the MSP database and NIST SD302 to create our ridge ground truth set. Our approach consists of three main steps:

\begin{enumerate}[noitemsep]
\item \textbf{Pre-enhancement:} This step reduces background noise and darkens ridge pixels, thus improving the subsequent enhancement using the method in \cite{tang2017FingerNet}. The techniques applied include: (i) CLAHE (Contrast Limited Adaptive Histogram Equalization), (ii) Non-local Means Denoising, and (iii) Adaptive Threshold.
\item \textbf{Ridge and minutiae enhancement:} We use a mixed Gabor enhancement convolutional model\footnotemark to enhance fingerprint ridges and minutiae. Each latent undergoes enhancement with: (i) the original latent image, (ii) the pre-enhanced image, and (iii) the pre-enhanced image with increased contrast. This step focuses on segmentation, ridge, and minutiae improvement and restoration. Segmentation prevents false ridge and minutiae formation in the background, and the enhancement results in sharper, noise-free ridges and minutiae. We fuse the prediction outputs to fill missing areas and reinforce the ridges in the three outputs.
\item \textbf{Oriented Gabor Filter \cite{hong1998fingerprint}:} The final step involves applying an Oriented Gabor Filter to: (i) binarize the latent print, generating the ground truth ridge masks for training and (ii) identify and eliminate unrecoverable corrupted regions within the fingerprint. The outcome is a ground truth binary latent mask suitable for training and addressing small corrupted regions.
\end{enumerate}

\footnotetext{\url{https://github.com/592692070/FingerNet/tree/master/models/released_version}}

\begin{figure}[H]
\centering
\renewcommand{\arraystretch}{0.5}
\setlength{\tabcolsep}{1pt}
\begin{tabular}{cccc}

\includegraphics[height=3.5cm]{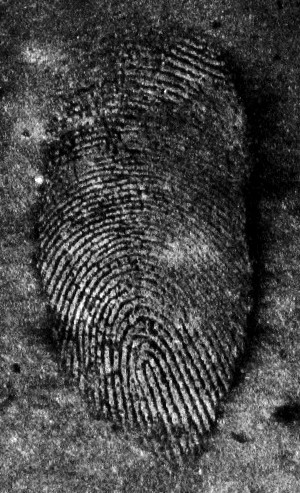}&
\includegraphics[height=3.5cm]{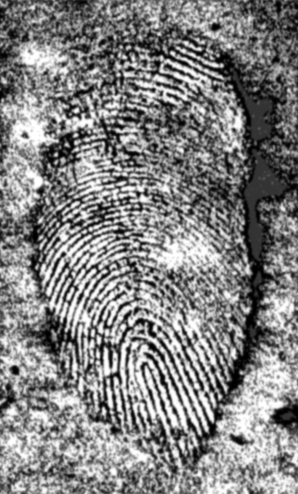}&
\includegraphics[height=3.5cm]{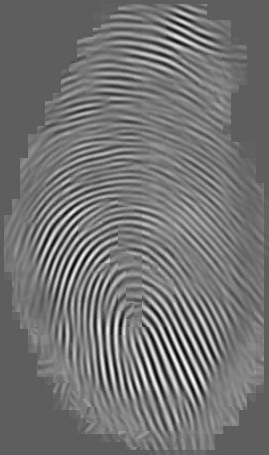}&
\includegraphics[height=3.5cm]{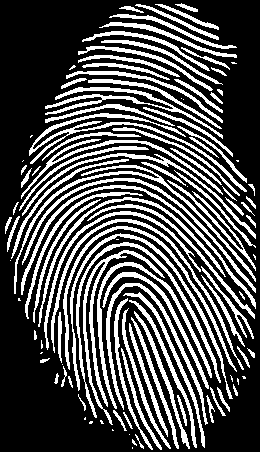}
\\
\scriptsize Latent fingerprint & \scriptsize Pre-enhancement & \scriptsize Ridge and minutiae & \scriptsize Oriented Gabor\newline \\ & & \scriptsize enhancement & \scriptsize filter

\end{tabular}
\caption{Training set creation steps. The result is a ground truth binary latent mask suitable for training.}
\label{traincreation}
\end{figure}

Although our method for creating the ground truth (see Figure \ref{traincreation} for our training set is designed for latents, it must be acknowledged that it can be time-consuming and not practical in real scenarios of forensic analysis. Consequently, our primary objective is to create a suitable ground truth training set, ultimately facilitating the creation of a segmentation model that excels in precision and speed.

\section{Proposed method}

Our latent fingerprint enhancement method, called ULPrint, comprises two primary steps (see Figure \ref{Flowchart}): (1) Ridge Segmentation, employing a UNet architecture with a SegFormer-B5 encoder \cite{DBLP:journals/corr/abs-2105-15203} for training, and (2) Final enhancement operations, involving a guided blending of the predicted ridge mask and latent image to maintain textures, ridge patterns, and other significant features. The outcome is an enhanced latent fingerprint image with increased clarity and distinct ridge structures suitable for deep network models utilizing image descriptors for recognition, such as MSU-AFIS \cite{cao2018end}, DeepPrint \cite{engelsma2019learning} or AFR-Net \cite{grosz2022afrnet}. Pseudocode \ref{alg:ridge_segmentation_enhancement} offers an overview of our approach, with additional details provided in the subsequent sections.

\begin{algorithm}
\caption{Latent Fingerprint Enhancement}\label{alg:ridge_segmentation_enhancement}
\begin{algorithmic}
\renewcommand{\algorithmicrequire}{\textbf{Input:}}
\Require Latent fingerprint image $latentImg$
\renewcommand{\algorithmicensure}{\textbf{Output:}}
\Ensure Enhanced fingerprint image: $enhancedImg$

\renewcommand{\algorithmicrequire}{\textbf{Step 1: Ridge Segmentation (UNet-SegFormer-B5):}}
\Require

\State \small a. Initialize UNet with SegFormer-B5 encoder
\State \small b. Apply multiple dilated convolutions
\State \small c. Training (Combined Dice loss and Focal loss $CLoss$, Adam)
\State \small d. Save the best model (validation IoU)
\State \small e. Segment ridges to obtain $ridgeMask$

\renewcommand{\algorithmicrequire}{\textbf{Step 2: Final enhancements operations:}}
\Require

\State \small $guidedL \gets GuidedBlend(ridgeMask, latentImg)$

\State \small $enhancedImg \gets Blend(latentImg, guidedL)$

\State \small  \textit{\textbf{output:} $enhancedImg$}
\end{algorithmic}
\end{algorithm}

\subsection{Segmentation with UNet and SegFormer-B5}
\label{unetresnet}

Segmentation of latent fingerprint ridges can help enhance the quality and accuracy of fingerprint recognition algorithms. In this section, we describe our approach to segmenting ridges using convolutional neural networks (CNNs) and Vision Transformer (ViT), adjusting the receptive field size of the network's convolutional filters to capture a larger context of the input image. Modulating the receptive field size of the network can enhance the ability to discriminate ridges from the background.

We implemented a U-Net architecture \cite{ronneberger2015u} with a Mix Visual Transformer (MiT) SegFormer-B5 encoder \cite{DBLP:journals/corr/abs-2105-15203} pre-trained on ImageNet \cite{deng2009imagenet}. By adding the Vision Transformer (ViT) to the architecture, it is possible to boost latent fingerprint quality further. ViT learns global context and detailed spatial relationships in images, making it suitable for ridge segmentation tasks.

The training utilized combined Dice loss function \cite{sudre2017generalised} with a Focal loss function \cite{DBLP:journals/corr/abs-1708-02002}. This combination can be advantageous for segmenting ridges and minutiae in fingerprints for several reasons: (i) Dice loss is particularly effective for segmentation tasks as it measures the overlap between the predicted and ground truth segmentation masks. It is less sensitive to class imbalances, which can be beneficial when dealing with fingerprint images where the background and ridge areas might differ significantly in size. (ii) Focal loss is designed to address the issue of class imbalance by down-weighting easy examples and focusing on hard latent instances. In fingerprint ridge and minutiae segmentation, distinguishing ridges, valleys, and the background might be challenging due to noise, distortions, or poor quality. Focal loss helps the model focus on these complex cases, potentially leading to better segmentation performance. We have named our combined loss function $CLoss$, which is expressed as:

{\small
\begin{equation}
CLoss = 0.5 \times (1 - \frac{2 |A \cap B|}{|A| + |B|}) - 0.5 \times \alpha (1 - p_t)^{\gamma} \log(p_t),
\end{equation}
}

Here, $A$ denotes the predicted mask, and $B$ represents the ground truth ridge mask. The Dice loss ranges from 0 to 1, where 0 indicates perfect overlap and 1 indicates no overlap. $p_t$ signifies the predicted probability of the true class, $\alpha$ is a balancing factor (between 0 and 1), and $\gamma$ is the focusing parameter that controls the down-weighting of easy examples. 

During the training, we applied random cropping to produce 256 $\times$ 256 patches from training images, encouraging the model to learn local features like ridge details and minutiae. Although trained on 256 $\times$ 256 patches, the fully convolutional U-Net can process arbitrary image sizes (512 $\times$ 512) or higher. This adaptability allows the model to accommodate diverse latent fingerprint databases with varying input dimensions.

Recognizing the inherent challenges of discerning small ridge structures in latent fingerprints, where the pixels of ridges and valleys are often indistinguishable, we opted to employ convolutional dilatation techniques \cite{yu2016multiscale} to augment the receptive field of the neural network. By altering the spacing between kernel elements, convolutional dilatation allows the network to perceive intricate, non-local patterns. In essence, dilated convolutions involve the introduction of a dilation factor, a parameter that expands the kernel's field of view without increasing its size or computational complexity. By modifying the dilation rates of the convolutional filters within the network's decoder blocks, we aimed to enhance the model's capacity to identify and segment ridge structures, capturing detailed texture information and handling varying ridge densities.

We adjusted the dilation rate to \textit{(1,1),(2, 2),(4,4)} across each of the four UNet decoder blocks, excluding the final one. This strategy facilitated the capture of more expansive image features from the latents, consequently boosting the network's ability to perform the segmentation. The specific dilatation operations employed throughout are listed within Table \ref{Architecture}, where they are highlighted. 

\begin{table}[H]
\centering
\resizebox{0.76\columnwidth}{!}{%
\begin{tabular}{@{}ll@{}}
\toprule
\textbf{Layer/Part}                                    & \textbf{Dilation rate} \\ \midrule
\textbf{Encoder: SegFormer-B5}    &                                             \\ \midrule
\textbf{Decoder: Unet}            &                                             \\
DecoderBlock: \textbf{(dilated)} & \multicolumn{1}{c}{(1,1), (2,2), (4, 4)}    \\
DecoderBlock: \textbf{(dilated) }& \multicolumn{1}{c}{(1,1), (2,2), (4, 4)}    \\
DecoderBlock: \textbf{(dilated) } & \multicolumn{1}{c}{(1,1), (2,2), (4, 4)}    \\
DecoderBlock: \textbf{(dilated)} & \multicolumn{1}{c}{(1,1), (2,2), (4, 4)}    \\
DecoderBlock                                      &                                             \\ \midrule
\textbf{Segmentation Head}&                                             \\
Conv2d and Softmax                                                                  &                                             \\ \bottomrule
\end{tabular}
}
\caption{Architecture of the solution}
\label{Architecture}
\footnotesize{* SegFormer-B5 has 12 blocks: multi-head self-attention, feedforward network, and depthwise separable convolution layer.}
\end{table}

Our training process involved training the model for 300 epochs and saving the best-performing model based on the validation IoU (Intersection over Union) score. Also, we used the Adam optimizer \cite{kingma2014adam}, and a learning rate of \textit{0.0001}. 

During the ridge mask prediction, we calculate the number of white pixels representing ridges and minutiae. If these white pixels make up less than 5\% of the input image's total pixels, we assume the model failed to segment ridges and minutiae. To address this, we apply the CLAHE technique \cite{pizer1987adaptive} to enhance local contrast, making fingerprint ridges more distinguishable from artifacts or noise. Next, we use the Fast Non-Local Means Denoising algorithm \cite{buades2011non} with a filter strength of \textit{5} for both luminance and chrominance components, a template patch window size of (3$\times$3), and a search window size of (7$\times$7), balancing noise reduction and preserving essential fingerprint features. We then blend the original and processed latent images using a weighted sum approach, assigning equal weights of \textit{0.5} to both images. This method ensures the latent image retains critical information while enhancing feature visibility. We rerun the model with the newly processed images, counting the white pixels again. If the count has increased, we save the new predicted mask.

\subsection{Augmentations during the training}
\label{aug}

We used various data augmentation techniques on the training images. These methods include:(i) Generating 256$\times$256 patches through random cropping to capture fine details of ridges and minutiae. (ii) Applying one of the following geometric transformations with a 75\% probability: horizontal flip, vertical flip, or random 90-degree rotation. (iii) Performing random resized cropping with a 50\% probability, adjusting the crop scale between \textit{0.8} and \textit{1.2}. (iv) Employing cutout with a 30\% probability, creating up to 5 holes with a maximum size of 10$\times$10 pixels each, simulating occlusions often found in latent fingerprints. Subsequently, we produced two models. One model included the following augmentations: (v) drawing random lines with a 30\% probability; (vi) drawing random letters with a 30\% probability. The other model did not have (v) and (vi) augmentations. We designated these models as $M_1$ and $M_2$, respectively.

\section{Fingerprint enhancement through guided blending}


We blend latent fingerprints with the predicted ridge mask from our UNet and SegFormer-B5 segmentation model (Section \ref{unetresnet}). Then, to generate the final enhanced latent fingerprint image, we employed the Guided Filter algorithm \cite{6319316}, which requires four input parameters: the latent input image, the guidance mask (predicted ridge latent mask), the filter radius $r = 5$, and a regularization term $\epsilon = 1 \times 10^{-6}$. The Guided Filter effectively preserves the edges of ridges and minutiae while reducing noise, enhancing details, and facilitating image fusion. The following equation can represent the algorithm:

\begin{equation}
q_i = \bar{a}_k \cdot I_i + \bar{b}_k
\end{equation}

where $q_i$ is the output enhanced latent fingerprint, $I_i$ is the input latent, and $\bar{a}_k$ and $\bar{b}_k$ are the local mean values of the linear coefficients $a$ and $b$ within a window of size $k$. The coefficients $a$ and $b$ are derived as follows:

\begin{align}
a = \frac{cov_{Ip}}{var_I + \epsilon}, \quad
b = \bar{p} - a \cdot \bar{I}
\end{align}

The Guided Filter algorithm utilizes several mathematical operations that help to improve the latent fingerprint quality. The covariance of the latent input image and the guidance predicted mask, represented by $cov_{Ip}$, measures how the latent input image and the guidance ridge mask are related. The variance of the latent image, denoted by $var_I$, provides insight into the dispersion of the pixel intensity values in the image. The regularization term $\epsilon$ is introduced to ensure stability during the calculations, while $\bar{I}$ and $\bar{p}$ represent the local mean values of the latent and guidance ridge mask, respectively.

The linear coefficients $a$ and $b$ are crucial in the enhancement process, as they are computed based on the derived covariance and variance of the input latent and the guidance ridge mask. The coefficient $a$ is obtained by dividing $cov_{Ip}$ by the sum of $var_I$ and $\epsilon$, which adjusts the relationship between the latent image and the guidance ridge mask. The coefficient $b$ is calculated by subtracting the product of $a$ and $\bar{I}$ from $\bar{p}$, emphasizing the contribution of the guidance mask in the enhancement. To account for the local context in the image, the local mean values of coefficients $a$ and $b$ are computed using a blurring function.

The final enhanced latent fingerprint image is generated by combining the input image with the mean coefficients $\bar{a}_k$ and $\bar{b}_k$. This operation preserves the ridge edges and minutiae while reducing noise, enhancing details, and facilitating image fusion. Algorithm \ref{algo} presents the complete pseudo-code for the guided latent fingerprint enhancement, illustrating the steps and objectives of the operations performed.

Finally, we blend the guided filtered latent fingerprint $q$ with the original latent fingerprint image using the weighted sum approach, assigning a $0.2$ weight to the latent image and a $0.8$ weight to the $q$ guided filtered latent fingerprint, where that is the outcome of our method. 

\begin{algorithm}[H]
\caption{Guided Filter for Latent Enhancement}\label{alg:guided_filter_enhancement}
\begin{algorithmic}
\renewcommand{\algorithmicrequire}{\textbf{Input:}}
\Require Latent image $I$, guidance ridge mask $p$, filter radius $r$, regularization term $\epsilon$
\renewcommand{\algorithmicensure}{\textbf{Output:}}
\Ensure Enhanced latent image: $q$

\renewcommand{\algorithmicrequire}{\textbf{Step 1: Compute local mean values:}}
\Require

\State \small a. Calculate local mean of $I$ ($\bar{I}$)
\State \small b. Calculate local mean of $p$ ($\bar{p}$)

\renewcommand{\algorithmicrequire}{\textbf{Step 2: Compute covariance and variance:}}
\Require

\State \small a. Compute covariance of $I$ and $p$ ($cov_{Ip}$)
\State \small b. Compute variance of $I$ ($var_I$)

\renewcommand{\algorithmicrequire}{\textbf{Step 3: Calculate linear coefficients:}}
\Require

\State \small a. Calculate $a = \frac{cov_{Ip}}{var_I + \epsilon}$
\State \small b. Calculate $b = \bar{p} - a \cdot \bar{I}$

\renewcommand{\algorithmicrequire}{\textbf{Step 4: Compute local mean values of coefficients:}}
\Require

\State \small a. Calculate local mean of $a$ within a window of size $k$ ($\bar{a}_k$)
\State \small b. Calculate local mean of $b$ within a window of size $k$ ($\bar{b}_k$)

\renewcommand{\algorithmicrequire}{\textbf{Step 5: Generate enhanced latent fingerprint image:}}
\Require

\State \small a. Compute $q = \bar{a}_k \cdot I + \bar{b}_k$

\State \small \textit{output: $q$ (guided filtered latent fingerprint)}
\end{algorithmic}
\label{algo}
\end{algorithm}

\begin{figure}[htb]
\centering
    \setlength{\tabcolsep}{3pt}
            \begin{tabular}{ccc}
            Input: latent &  Guided filter & Enhanced latent \\
            \includegraphics[height=2.6cm]{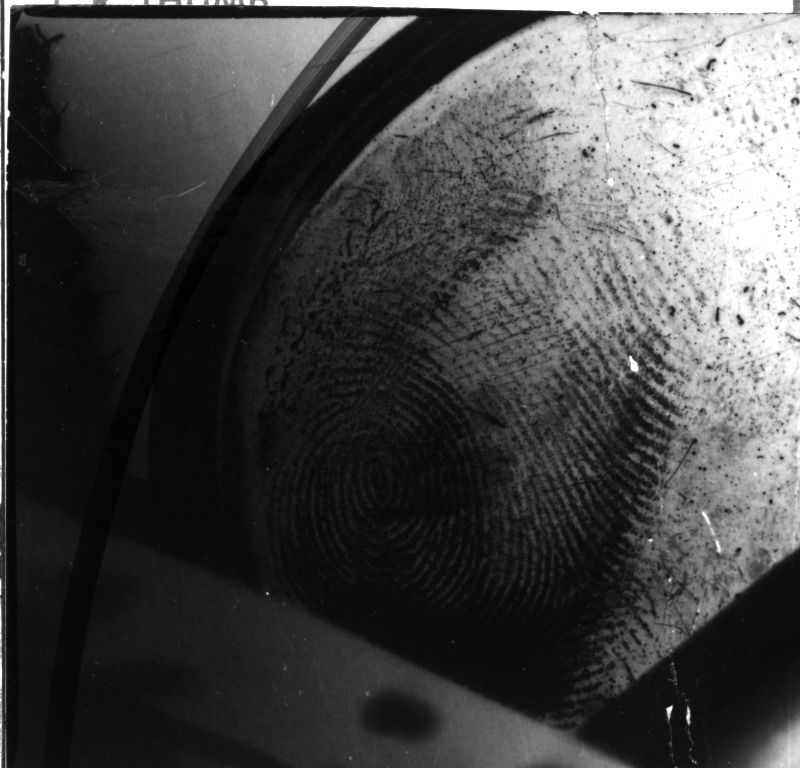}&
          \frame{ \includegraphics[height=2.6cm]{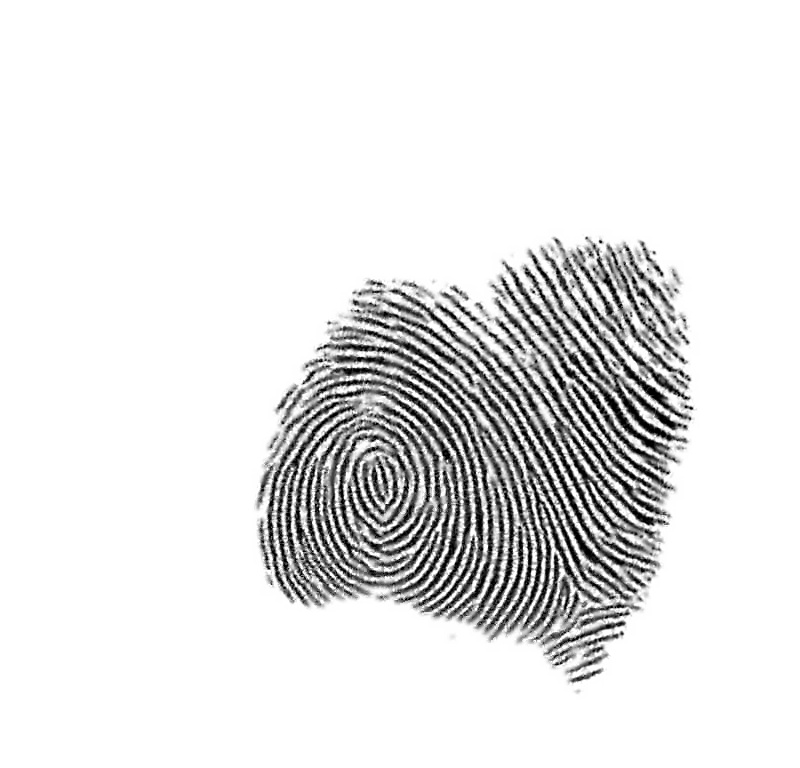}}&
            \includegraphics[height=2.6cm]{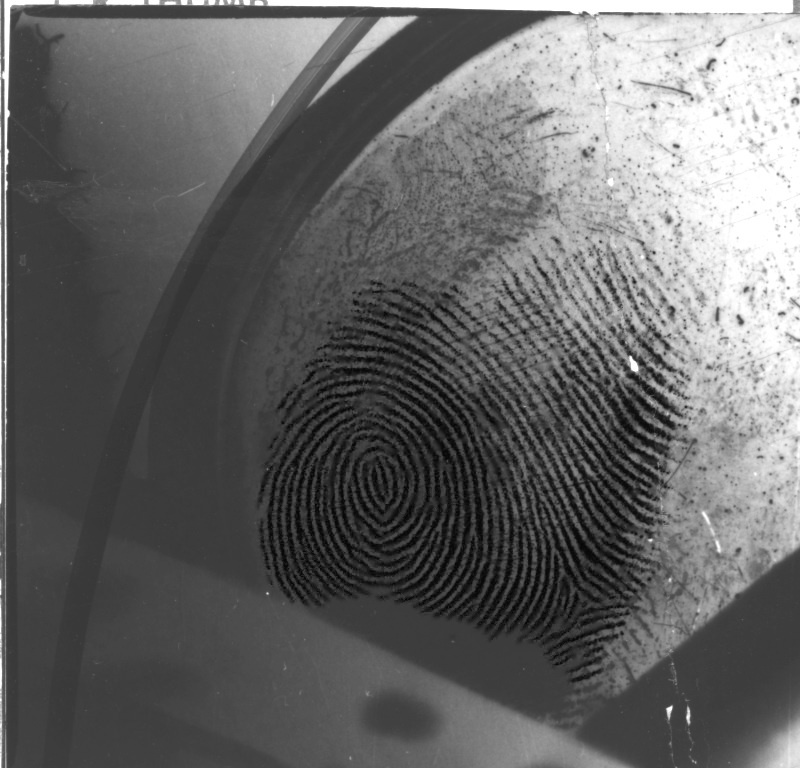}

            \end{tabular}
            \caption{Latent fingerprint enhancement process}
            \label{sd27examples}
\end{figure}

\section{Experimental Results}

\subsection{Implementation details}

Our model is built using PyTorch and Python, incorporating the Segmentation Models PyTorch API\footnote{Segmentation Models PyTorch API: \url{https://smp.readthedocs.io/en/latest/}}. We employ a parallelization strategy to optimize computational efficiency that leverages multi-GPUs and processors, accelerating the inference process. This approach allows for the simultaneous processing of multiple latent fingerprint images, reducing the overall inference time of our model.

\section{Qualitative results - Visual comparison}

This section presents our study's qualitative results by visually comparing the binarization outcomes and the final texture enhancements achieved using various methods, including our UNet and SegFormer-B5-based model. Our goal is to demonstrate the efficacy of our approach in producing better-quality latent fingerprint images that retain essential features necessary for accurate fingerprint recognition.

To initiate the visual comparison, we focus on the binarization results of different methods, comparing them against our UNet and SegFormer-B5-based model. This step is crucial as alternative techniques, such as the Oriented Gabor Filter \cite{hong1998fingerprint} and Verifinger SDK V12.4, cannot seamlessly integrate and blend the mask with the original latent fingerprint image. By contrasting the binarization outcomes, we aim to underscore the superior performance of our model.

\begin{figure}[H]
\centering
    \setlength{\tabcolsep}{1pt}
            \begin{tabular}{ccc}

            \footnotesize NIST SD27 & \footnotesize MSP Latent & \footnotesize NIST 302\\

            \includegraphics[height=2.6cm]{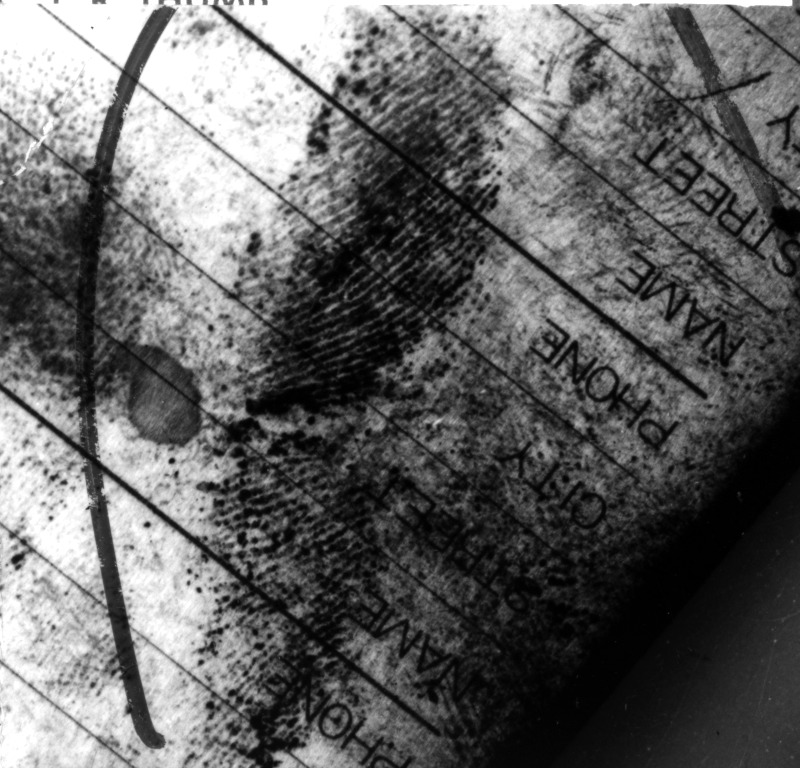}&
            \frame{\includegraphics[height=2.6cm]{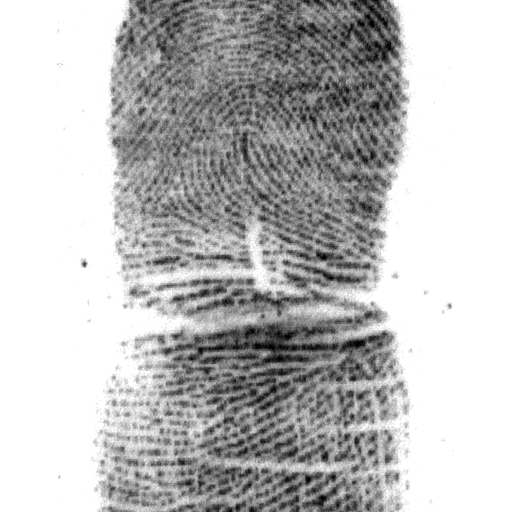}}&
            \frame{\includegraphics[height=2.6cm]{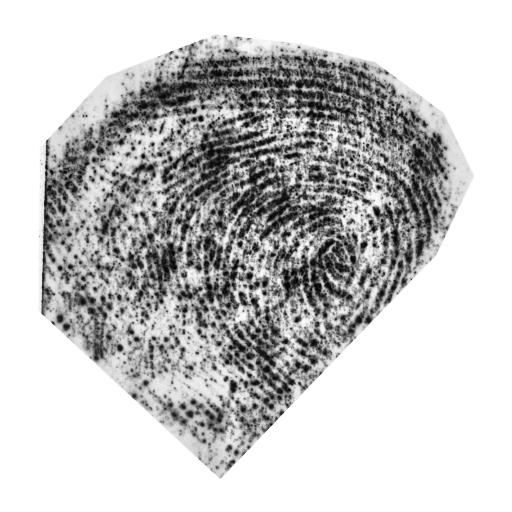}}\\

            \includegraphics[height=2.6cm]{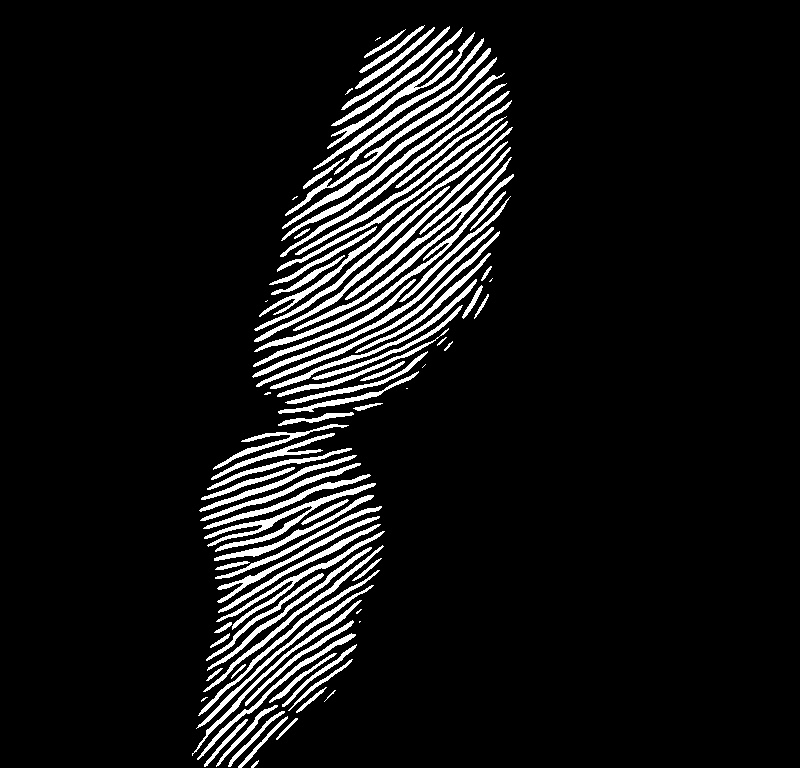}&
            \includegraphics[height=2.6cm]{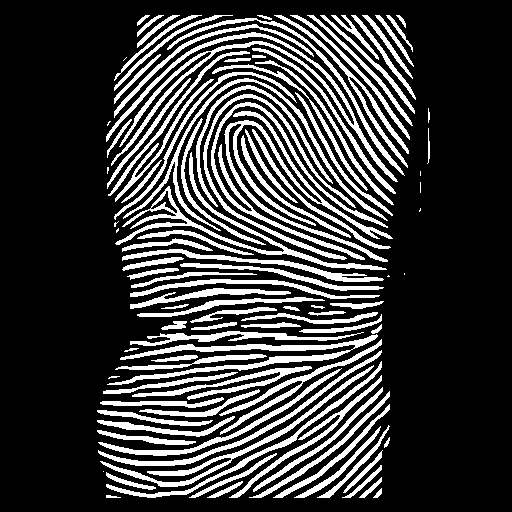}&
            \includegraphics[height=2.6cm]{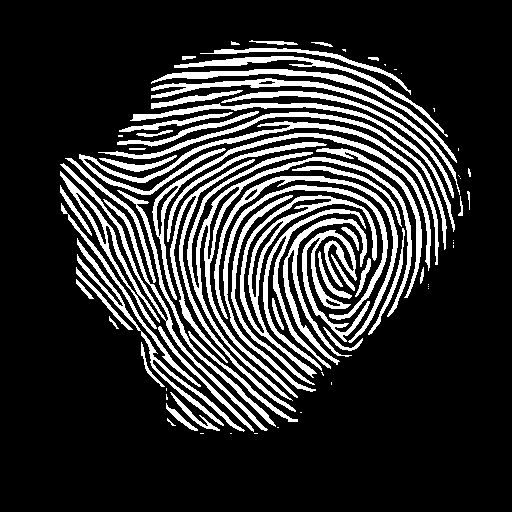}

            \end{tabular}
            \caption{Examples of latent images from the NIST SD27 database and their respective binary image prediction of ridges and minutiae segmented by our model. }
            \label{examples_latents}
\end{figure}

\begin{figure*}[!h]
	\centering
    \setlength{\tabcolsep}{1.2pt}
	\begin{tabular}{cccccc}
	    \multicolumn{2}{c}{NIST SD27} & \multicolumn{2}{c}{MSP Latent} & \multicolumn{2}{c}{NIST 302} \\
	    \begin{turn}{90}\tab[0.55cm]\footnotesize \hspace{-0.4cm} Latent image\end{turn}\hspace{0.05cm}

            \includegraphics[height=2.71cm]{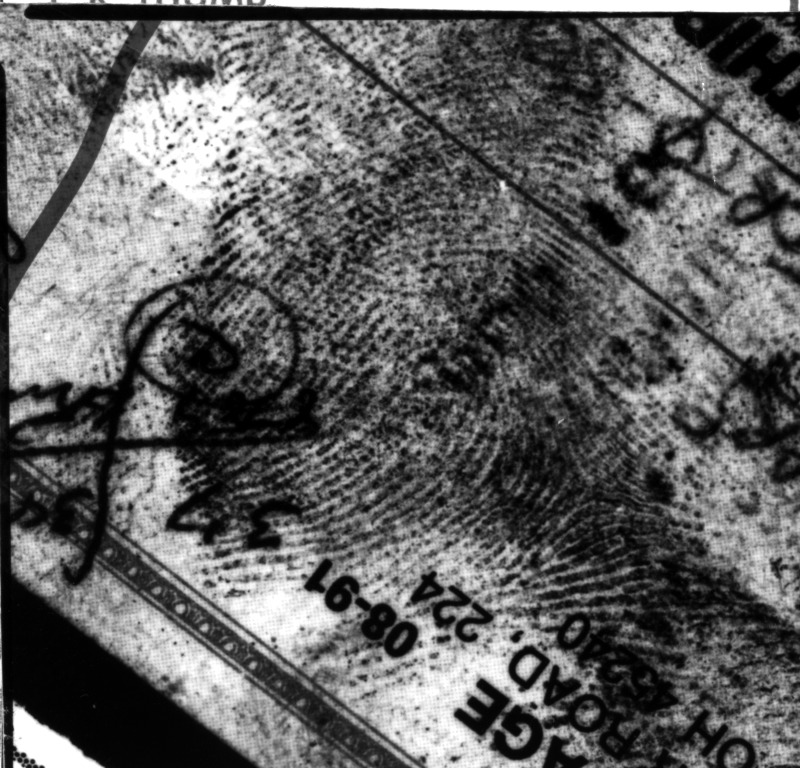} &   
	    \includegraphics[height=2.71cm]{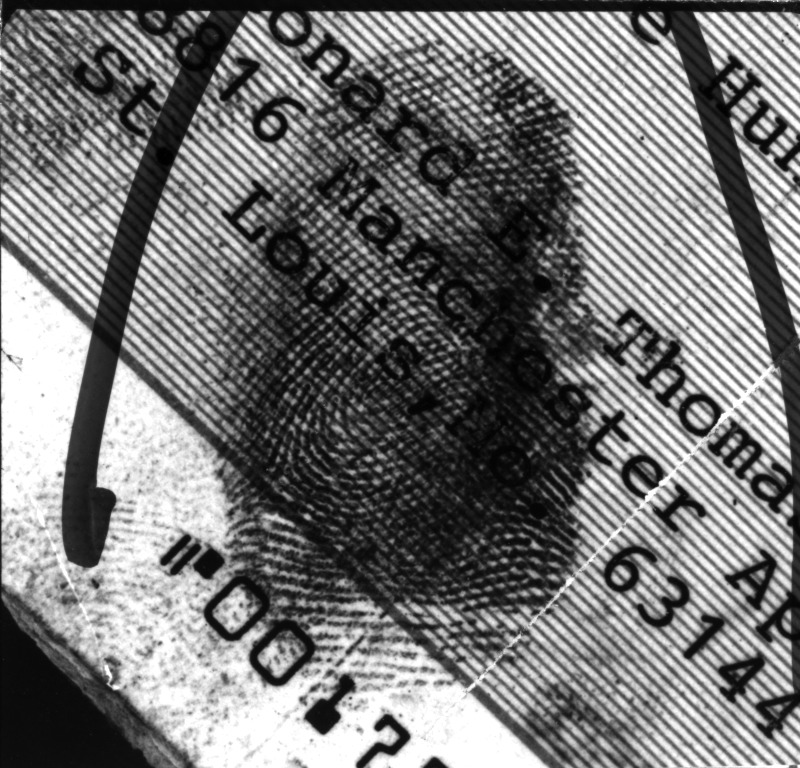} &   
	   \frame{ \includegraphics[height=2.71cm]{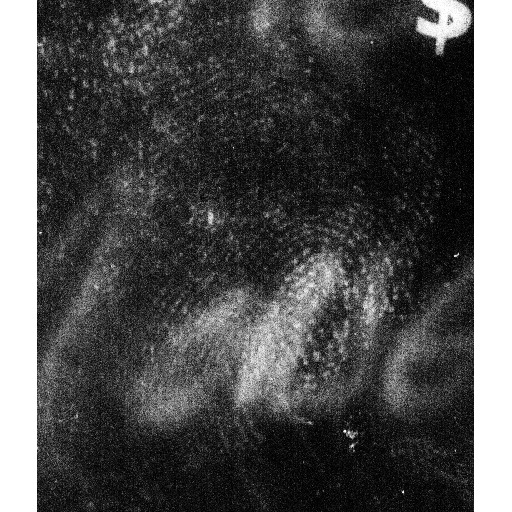}} &   
	   \frame{ \includegraphics[height=2.71cm]{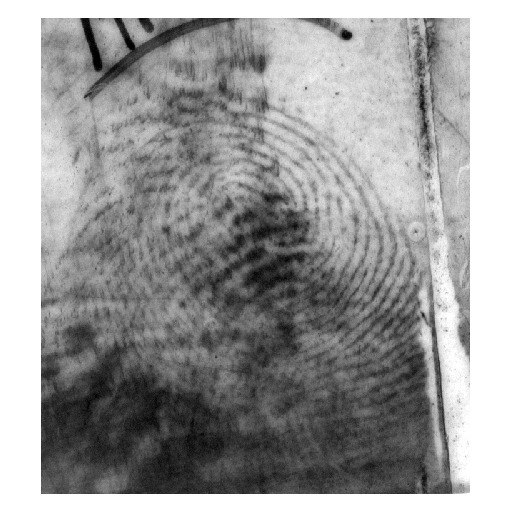}} &   
	   \frame{ \includegraphics[height=2.71cm]{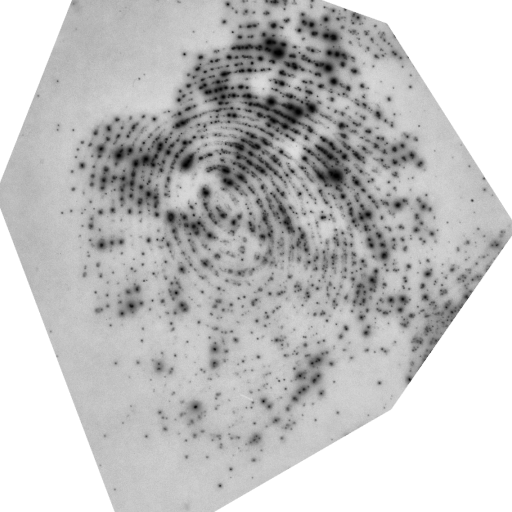} }&   
	   \frame{ \includegraphics[height=2.71cm]{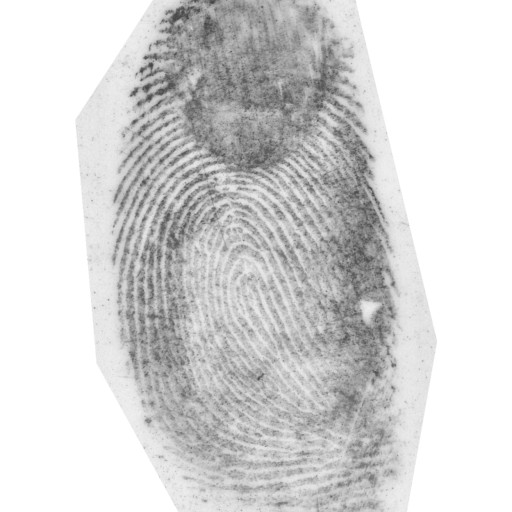}}\\
     
     \vspace{-0.2cm}\\ \hline \vspace{-0.1cm}\\

		\begin{turn}{90}\tab[0.55cm]\footnotesize \hspace{0.0cm} ULPrint \end{turn}\hspace{0.05cm}\hspace{0.30cm}\includegraphics[height=2.71cm]{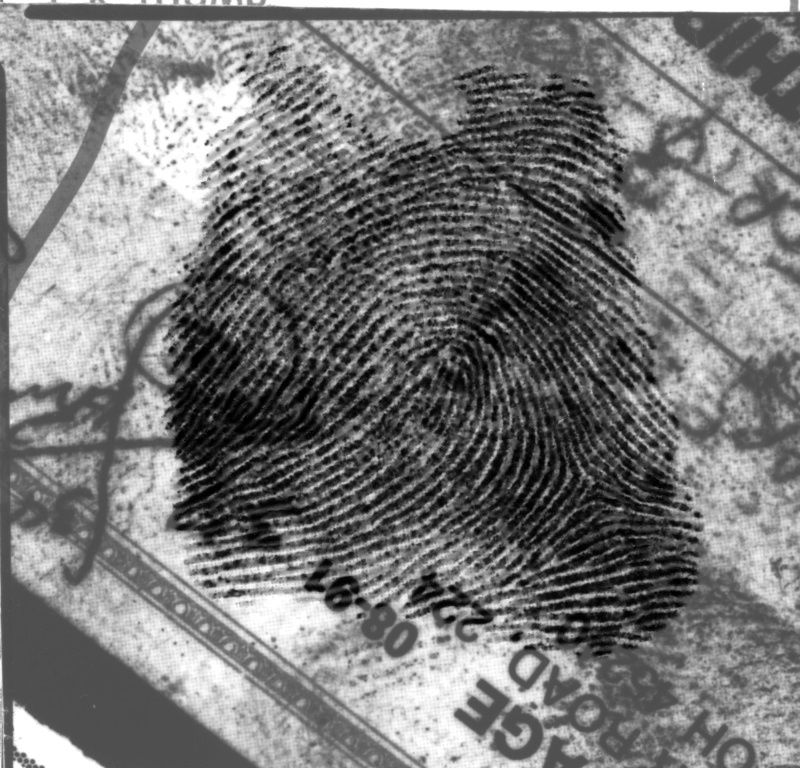} &   
		\includegraphics[height=2.71cm]{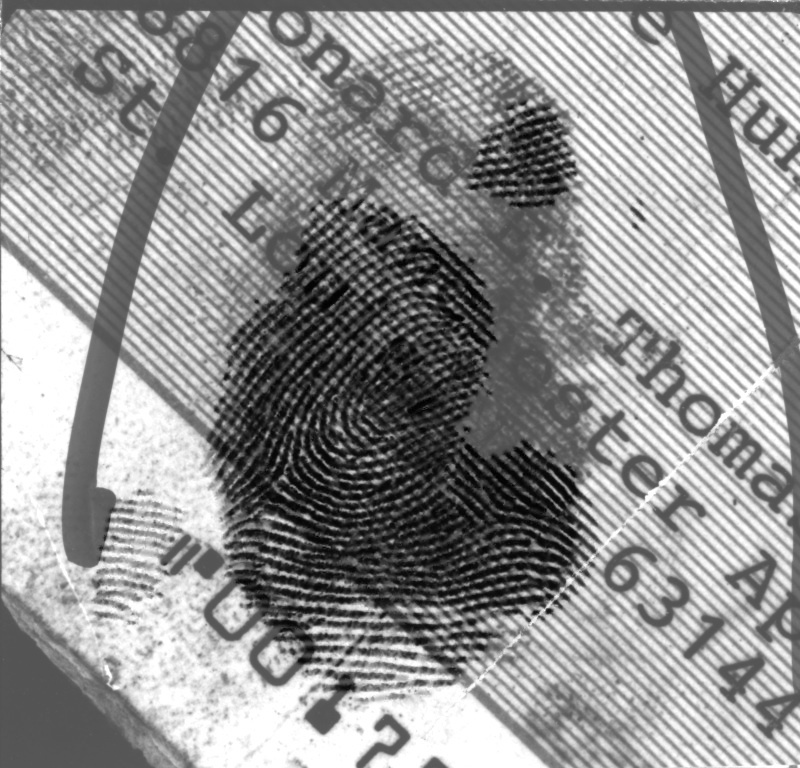} &   
		 \includegraphics[height=2.71cm]{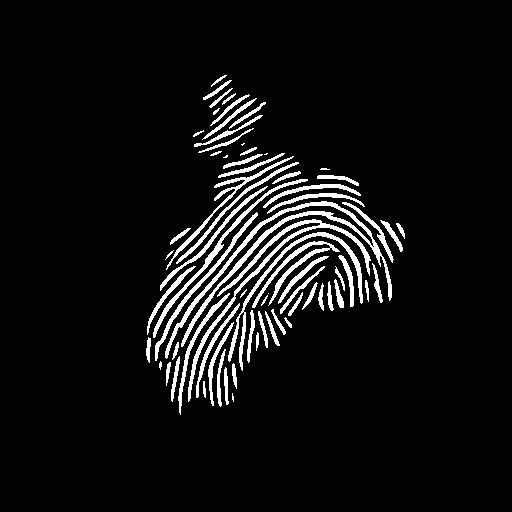} &   
		\includegraphics[height=2.71cm] {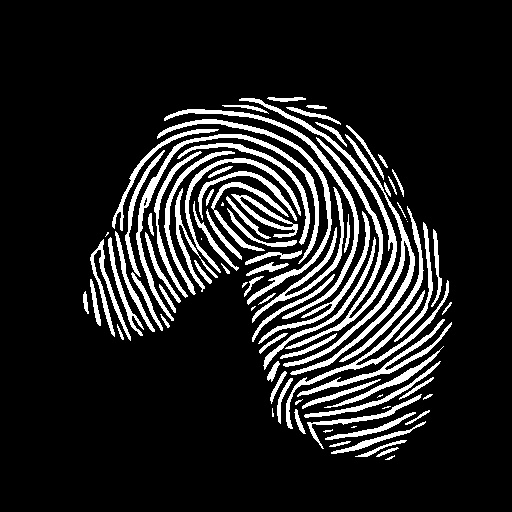} &   
		 \frame{\includegraphics[height=2.71cm]{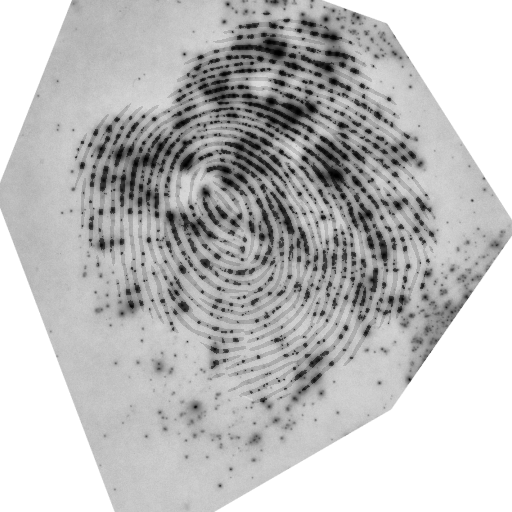}} &   
		 \frame{\includegraphics[height=2.71cm]{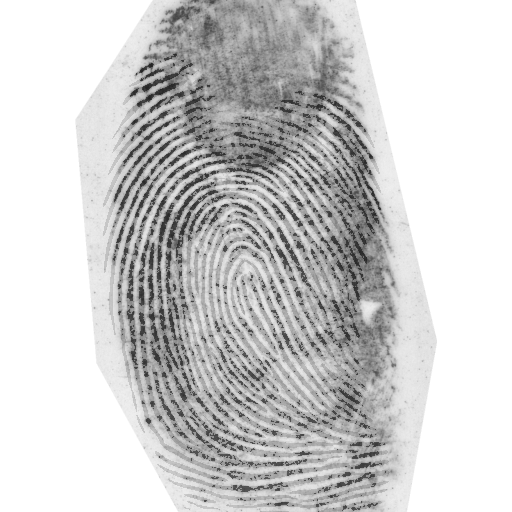}}\\ 
  
           \begin{turn}{90}\tab[0.55cm]\footnotesize \hspace{-0.4cm} ULPrint binary\end{turn}\hspace{0.05cm}
            \includegraphics[height=2.71cm]{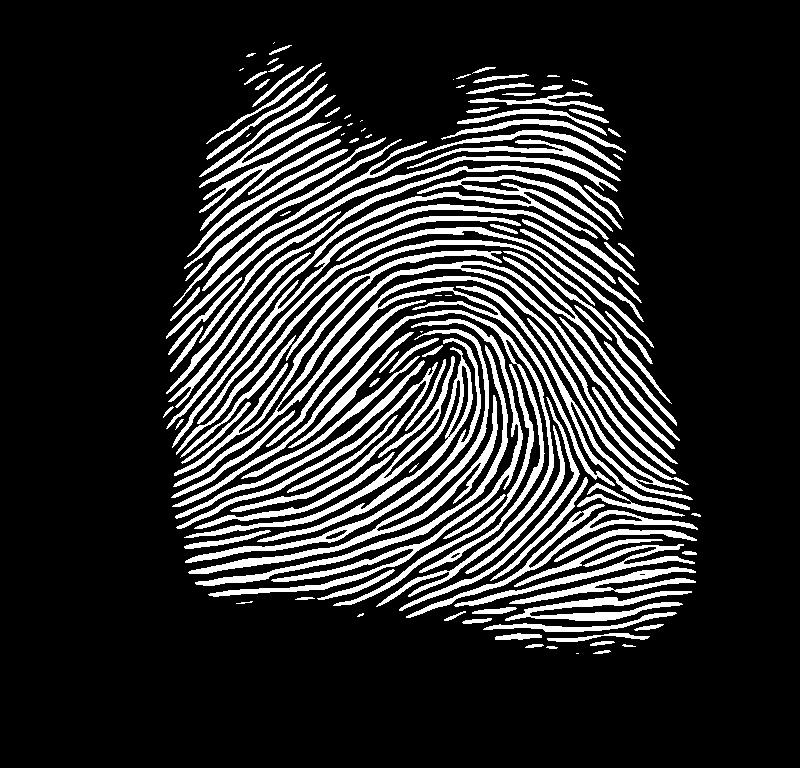} &   
	    \includegraphics[height=2.71cm]{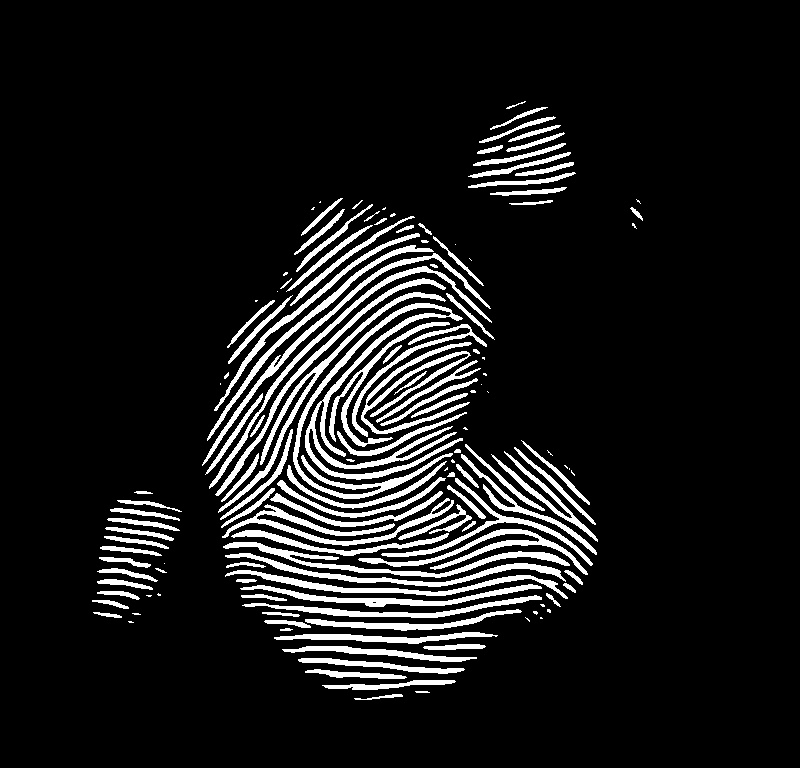} &   
	   \frame{ \includegraphics[height=2.71cm]{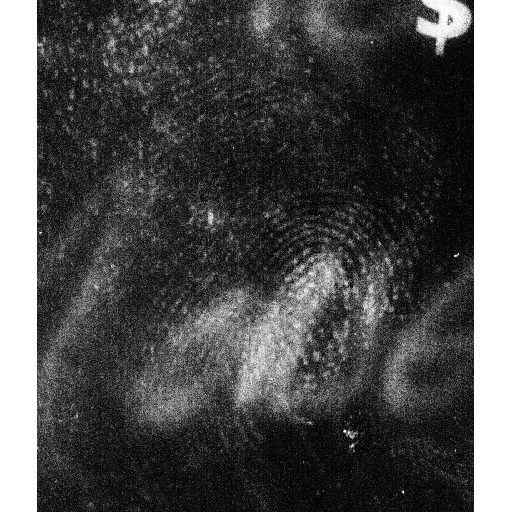} }&   
		\frame{\includegraphics[height=2.71cm]{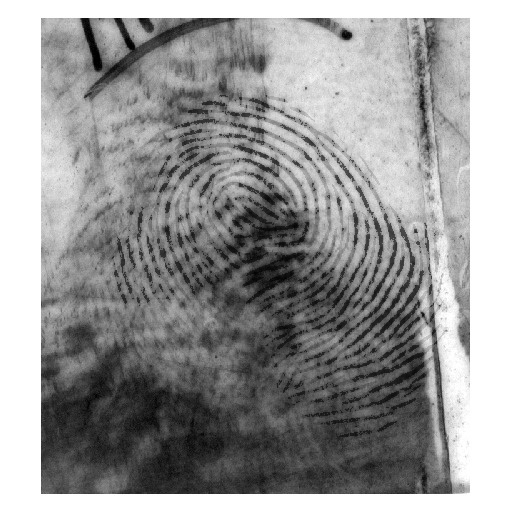} }&   
	    \includegraphics[height=2.71cm]{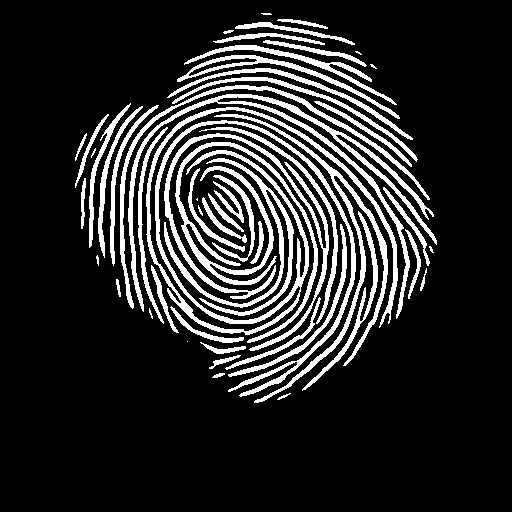} &   
	    \includegraphics[height=2.71cm]{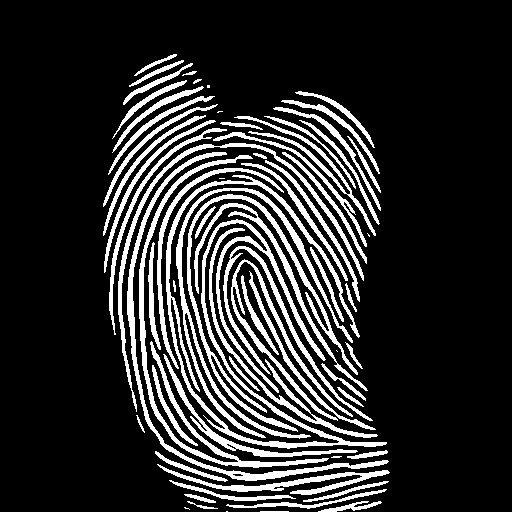}\\
     
  \vspace{-0.2cm}\\ \hline \vspace{-0.1cm}\\ 

  \begin{turn}{90}\tab[0.55cm]\footnotesize \hspace{-0.4cm} FingerNet \cite{tang2017FingerNet} \end{turn}
		\hspace{0.32cm}\includegraphics[height=2.71cm]{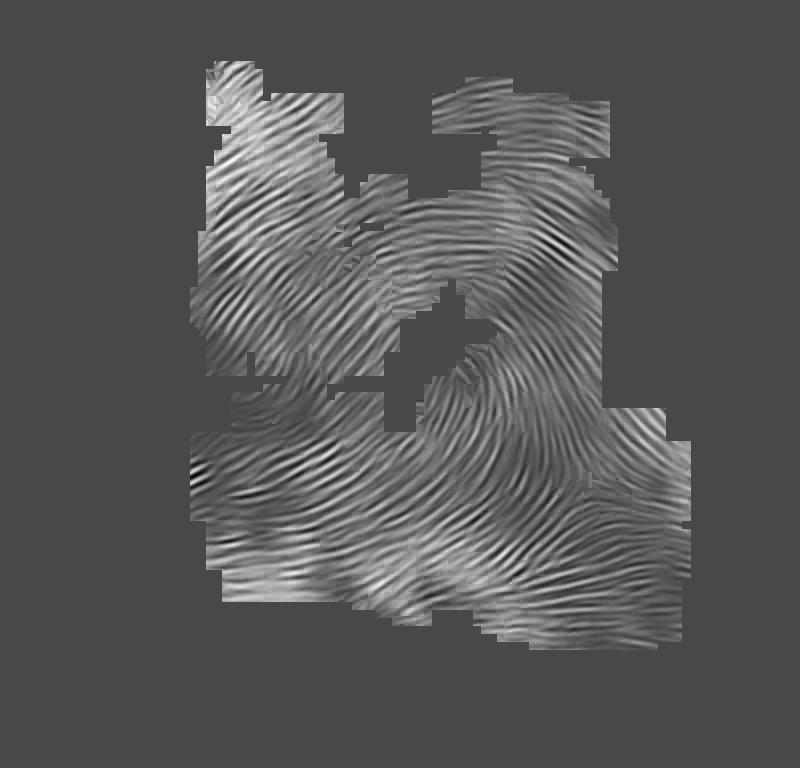} &   
		\includegraphics[height=2.71cm]{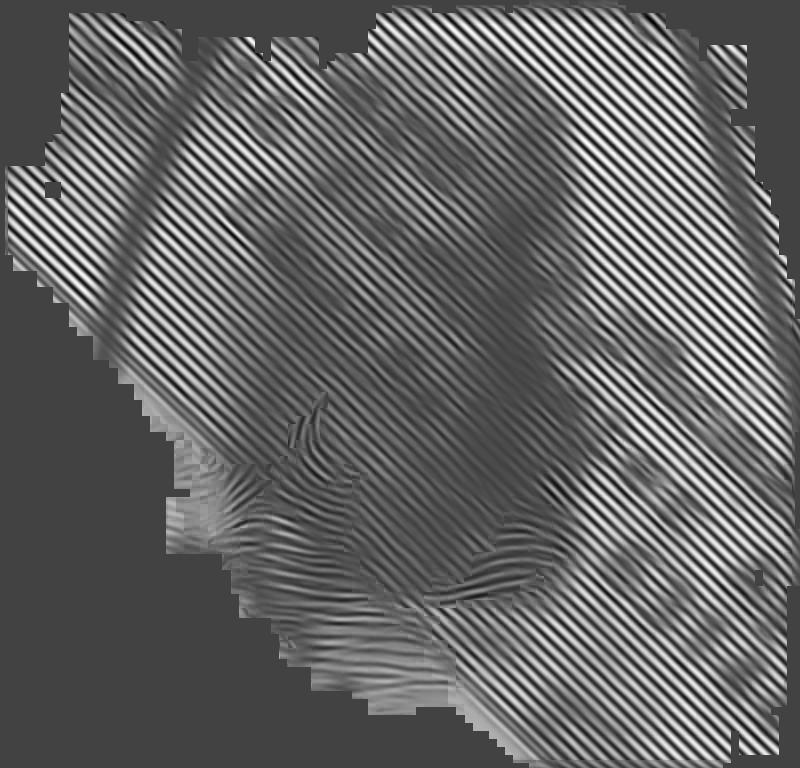} &   
		\includegraphics[height=2.71cm]{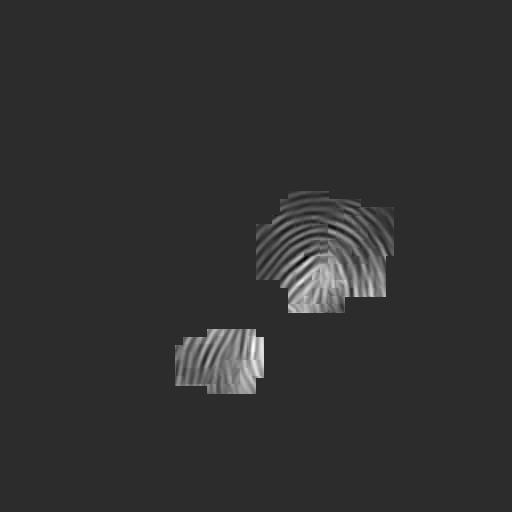} &   
		\includegraphics[height=2.71cm]{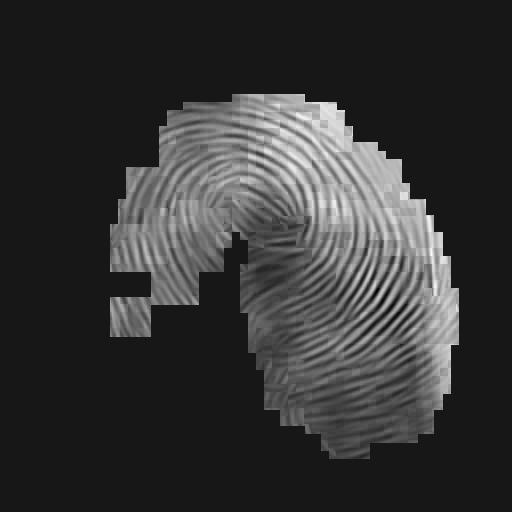} &   
		\includegraphics[height=2.71cm]{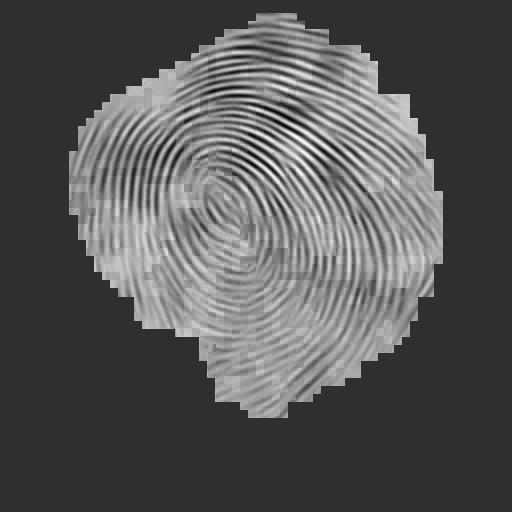} &   
		\includegraphics[height=2.71cm]{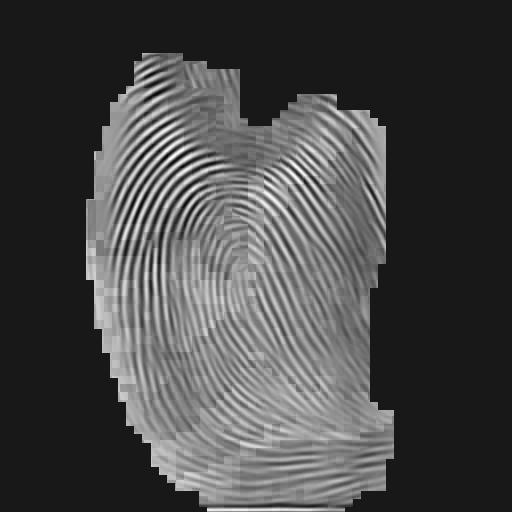}

	\end{tabular}
    \caption{Visual comparison of latents from NIST SD27, MSP Latent, and NIST 302, alongside their respective enhancements. These enhancements include our guided blended image (termed as ULPrint enhancement), our predicted ridges and minutiae (designated as ULPrint binary), and the images enhanced via FingerNet \cite{tang2017FingerNet}.}.
	\label{variation_backgroundNEW}
\end{figure*}

Subsequently, we examine the final texture enhancements generated by our approach (with the background removed), the Fingernet \cite{tang2017FingerNet} processing, and the enhancement provided by the MSU-AFIS autoencoder. It is important to note that although binarization plays a critical role in fingerprint image processing, new matching algorithms rely heavily on image features derived from ridge textures, minutiae, and other related characteristics. Figures \ref{examples_latents} and \ref{variation_backgroundNEW} display our predicted ridges and minutiae, which are designated as ULPrint binary. They also provide a visual comparison of latents from NIST SD27, MSP Latent, and NIST 302, together with their respective enhancements. Additionally, these figures include a comparison with images enhanced via FingerNet \cite{tang2017FingerNet}.

\subsection{Quality experimental results - NFIQ}

\begin{figure}[h]
\centering
            \includegraphics[height=3.84cm]{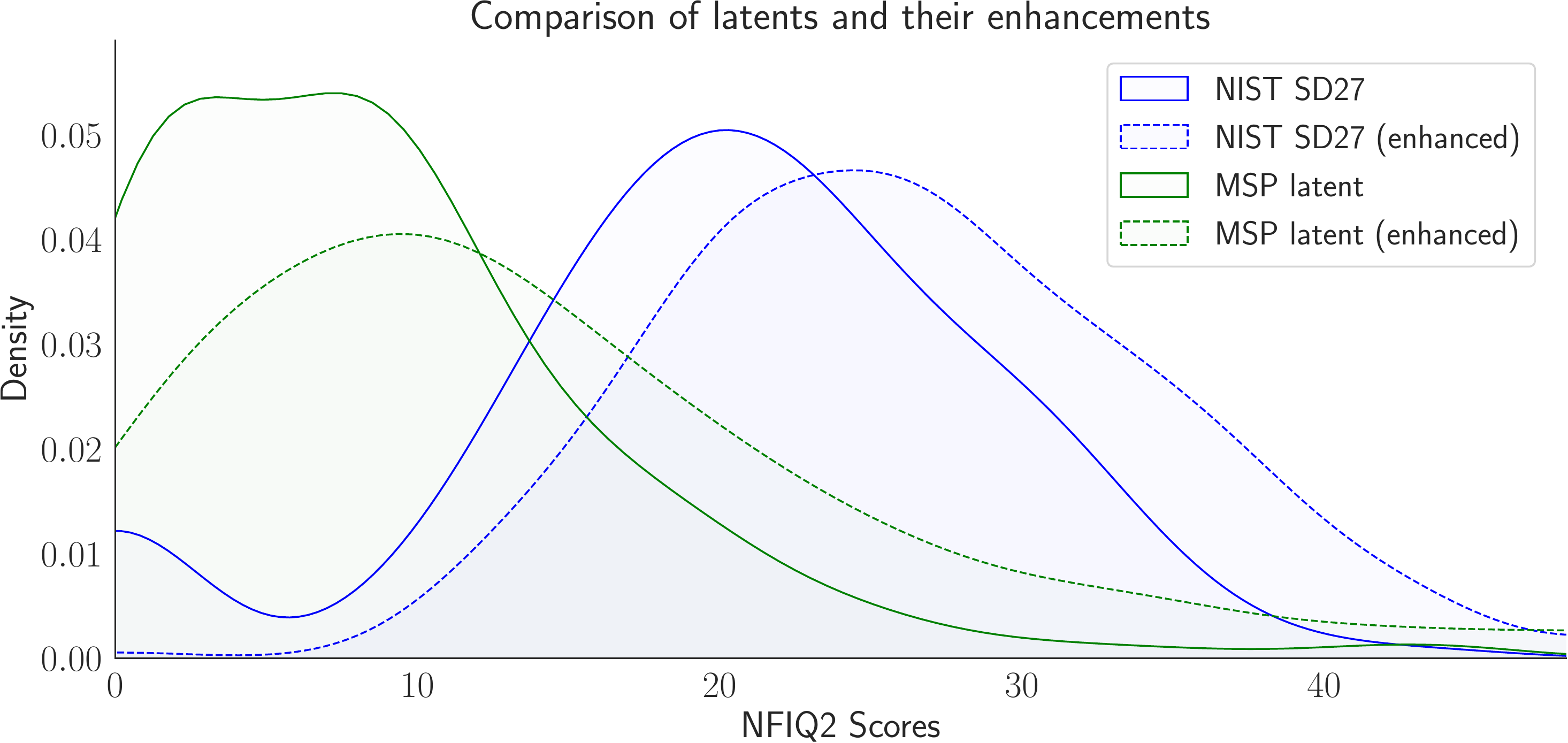}
            \caption{Histogram of NFIQ 2 scores for NIST SD27 and MSP Latent. Note that the enhancement by our method overlaps the quality histograms of the original latent images.}
            \label{nfiq2}
\end{figure}

In this experiment, our primary objective is to assess the quality of enhancement images generated by our method and compare them to their original images. To accomplish this, we utilize the evaluation set to evaluate the quality of the latent fingerprints and the enhanced latents. This analysis employs the NFIQ 2 metric \cite{bausinger2011fingerprint} as a means of quantification. The findings from our experiment (see Figure \ref{nfiq2} and Table \ref{nfiq2table}) indicate a significant improvement in image quality using our enhancement method. 

\begin{table}[h]
\centering
\resizebox{\columnwidth}{!}{%

\begin{tabular}{@{}lcccccc@{}}
\toprule
\textbf{Database} & \multicolumn{3}{c}{\textbf{Before enhancement}} & \multicolumn{3}{c}{\textbf{After enhancement}} \\ 
\cmidrule(lr){2-4} \cmidrule(lr){5-7}
 & Mean & Std Dev & Skewness & Mean & Std Dev & Skewness \\ \midrule
NIST SD27 & 20.04 & 8.88 & -0.54 & \textbf{26.62} & 8.42 & 0.46 \\
MSP Latent & 9.47 & 10.57 & 3.19 & \textbf{15.91} & 14.35 & 1.82 \\
NIST 302 & 4.67 & 7.88 & 3.13 & \textbf{16.31} & 16.02 & 1.17 \\
\bottomrule
\end{tabular}
}
\caption{Mean, standard deviation, and skewness of NFIQ 2 scores before and after enhancement.}
\label{nfiq2table}
\end{table}

\subsection{Identification (1:N Comparison)}

This experiment assesses identification accuracy through a closed-set analysis by matching latent prints from the NIST SD27, NIST 302 \cite{fiumara2019nist}, and MSP Latent databases against a gallery comprising 100,000 rolled fingerprints from both the MSP Database and NIST SD14 \cite{watson2001nist}. Our analysis results can be found in Table \ref{table_1_n}. We follow the evaluation protocol established in NIST ELFT-EFS \cite{flanagan2010nist}. This experiment used the unmodified MSU-AFIS matching method and the matching algorithm for template extraction.

\begin{table}[H]
\centering

\resizebox{0.46\textwidth}{!}{%

\begin{tabular}{ccccc}
\hline
\textbf{Model}    & \textbf{NIST SD27} & \textbf{MSP Latent} & \textbf{NIST SD302*}  \\ \hline
MSU-AFIS  & 61.56  & 67.63  & 46.90              \\
MSU-AFIS$_1$     & 73.64  & 76.84  & 50.38             \\
MSU-AFIS$_2$     & 74.41  & 76.52  & 51.23             \\
MSU-AFIS$_3$     & \textbf{75.19}  & \textbf{77.02}  & \textbf{52.12}            \\
Verifinger V12.4     & 27.51   &  55.06  & 52.08  \\ 
\hline
\end{tabular}
}
\\ \scriptsize * We reduced the resolution of N2N latent images to 500 dpi.
\caption{Rank-1 accuracies (\%) of different models against a background set of 100k rolled fingerprints, in addition to the true rolled mates of the latents prints of each database.}
\label{table_1_n}
\end{table}

\section{Conclusion}

We introduced a novel approach that uses deep learning and computer vision techniques to address the challenge of latent fingerprint enhancement, with the ultimate goal of improving the accuracy of fingerprint recognition systems. This method is structured as a two-step process for latent fingerprint segmentation and enhancement. It uses Ridge Segmentation in combination with the UNet and Mix Visual Transformer (MiT) SegFormer-B5 encoder architecture. Additionally, we integrated multiple dilated convolutions within the UNet architecture. This inclusion is designed to capture intricate, non-local patterns more effectively, thus improving the quality of ridge segmentation.

We validated the effectiveness of the proposed approach through a comprehensive series of experimental analyses. These included a visual comparison to illustrate enhancement quality, a Fingerprint Image Quality (NFIQ) assessment from the National Institute of Standards and Technology for quantitative quality analysis, and a closed-set identification accuracy analysis involving latent prints from the NIST SD27, NIST 302, and MSP Latent databases. In the closed-set identification accuracy experiments, the enhanced image was able to improve the performance of the MSU-AFIS from 61.56\% to 75.19\% in the NIST SD27 database, from 67.63\% to 77.02\% in the MSP Latent database, and from 46.90\% to 52.12\% in the NIST SD302 database. These results outperformed those achieved with the well-established Verifinger SDK 12.4. These tests showcased the method's capability to manage a wide variety of latent fingerprint types, highlighting its potential applicability in forensic investigations.

Our study also proposes a method for guided blending of the predicted ridge mask with the latent fingerprint. This technique not only optimizes the enhancement process but also enables the system to effectively deal with a broad range of latent fingerprint types.

The results provide a solid basis for further exploration in this field, promising to enhance latent fingerprint identification's precision and efficiency. Future research in this field could lead to even more advanced methods and applications within the broader scope of forensic science.



\normalsize
\bibliography{references}


\end{document}